\documentclass[11pt,a4paper]{article}
\usepackage[hyperref]{emnlp-ijcnlp-2019}
\usepackage{times}
\usepackage{latexsym}

\usepackage{url}

\usepackage{graphicx}
\usepackage{epstopdf}
\usepackage{epsfig}
\usepackage{amssymb}
\usepackage[normalem]{ulem}
\usepackage{url}
\usepackage{float}
\usepackage{multirow}
\usepackage{colortbl}
\usepackage{booktabs}
\usepackage{bm}
\usepackage{bbm}
\usepackage{amsmath}

\usepackage{array}
\usepackage{soul}
\usepackage{url}
\usepackage{graphicx}
\usepackage{amsmath}
\usepackage{multirow}
\usepackage{subfigure}
\usepackage{booktabs}
\usepackage{amssymb}
\usepackage{amsfonts}
\usepackage{algorithm}
\usepackage{algorithmic}
\usepackage{url}
\usepackage{bm}
\usepackage{enumitem}

\aclfinalcopy % Uncomment this line for the final submission

\title{Pun-GAN: Generative Adversarial Network for Generating Pun Text}
\title{Improving Neural Pun Generation with Generative Adversarial Nets}
\title{Improving Neural Pun Generation via Adversarial Learning}
\title{Adversarial Neural Pun Generation}
\title{Pun-GAN: Generative Adversarial Network for Pun Generation}
  
\author{
Fuli Luo\textsuperscript{1}\thanks{\ \ Equal Contribution.},
Shunyao Li\textsuperscript{1}\footnotemark[1],
Pengcheng Yang\textsuperscript{1},
Lei li\textsuperscript{1}, \\
\textbf{
Baobao Chang\textsuperscript{1,2},
Zhifang Sui\textsuperscript{1,2},
Xu Sun\textsuperscript{1}}\\
\textsuperscript{1}Key Lab of Computational Linguistics, Peking University\\
\textsuperscript{2}Peng Cheng Laboratory, China\\
\{luofuli, lishunyao, yang\_pc, lilei\_nlp, chbb, szf, xusun\}@pku.edu.cn
}

\date{}

\begin{document}
\maketitle
\begin{abstract}
  
\end{abstract}
% Pun generation is an interesting and challenging text generation task.
In this paper, we focus on the task of generating a pun sentence given a pair of word senses.
A major challenge for pun generation is the lack of large-scale pun corpus to guide the supervised learning.
To remedy this, we propose an adversarial generative network for pun generation (Pun-GAN), which does not require any pun corpus.
It consists of a generator to produce pun sentences, and a discriminator to distinguish between the generated pun sentences and the real sentences with specific word senses.
The output of the discriminator is then used as a reward to train the generator via reinforcement learning, encouraging it to produce pun sentences which can support two word senses simultaneously.
Experiments show that the proposed Pun-GAN can generate sentences that are more ambiguous and diverse in both automatic and human evaluation.\footnote{The code is available at: \url{https://github.com/lishunyao97/Pun-GAN}.}
% , combining generation and discrimination process via a mini-max game.
% The two sub models play a mini-max game and finally achieve the double win.

\section{Introduction}
Generating creative and interesting text is a key step towards building an intelligent natural language generation system.
A pun is a clever and amusing use of a word with two meanings (word senses), or of words with the same sound but different meanings~\cite{MillerG15}.
In this paper, we focus on the former type of pun, i.e., homographic pun.
For example, ``I used to be a banker but I lost \textit{\underline{interest}}'' is a pun sentence because the pun word ``\textit{\underline{interest}}'' can be interpreted as either \textit{curiosity} or \textit{profits}.

An intractable problem for pun generation is the lack of a large-scale pun corpus in which each pun sentence is labeled with two word senses.
Early researches~\cite{Hong2009,pun13_template,pun13_template2} are mainly based on templates and rules, thus lacking creativity and flexibility.
\citet{Neural18Yu} is the first endeavor to apply neural network to this task, which adopts a constrained neural language model~\cite{ConstrainedLM} to guarantee that a pre-given word sense to appear in the generated sequence.
However, \citet{Neural18Yu} only integrates the generation probabilities of two word senses during the \textit{inference} decoding process, without \textit{\textbf{detecting}} whether the generated sentences can support the two senses indeed during \textit{training}.
Promisingly, Word Sense Disambiguate (WSD)~\cite{WSD_survey} which aims at identifying the correct meaning of the word in a sentence via a multi-class classifier, can help the detection of pun sentences to some extent.

\begin{figure}[t]
	\centering
	%\vspace{-0.1in}
    \includegraphics[width=1.0\columnwidth]{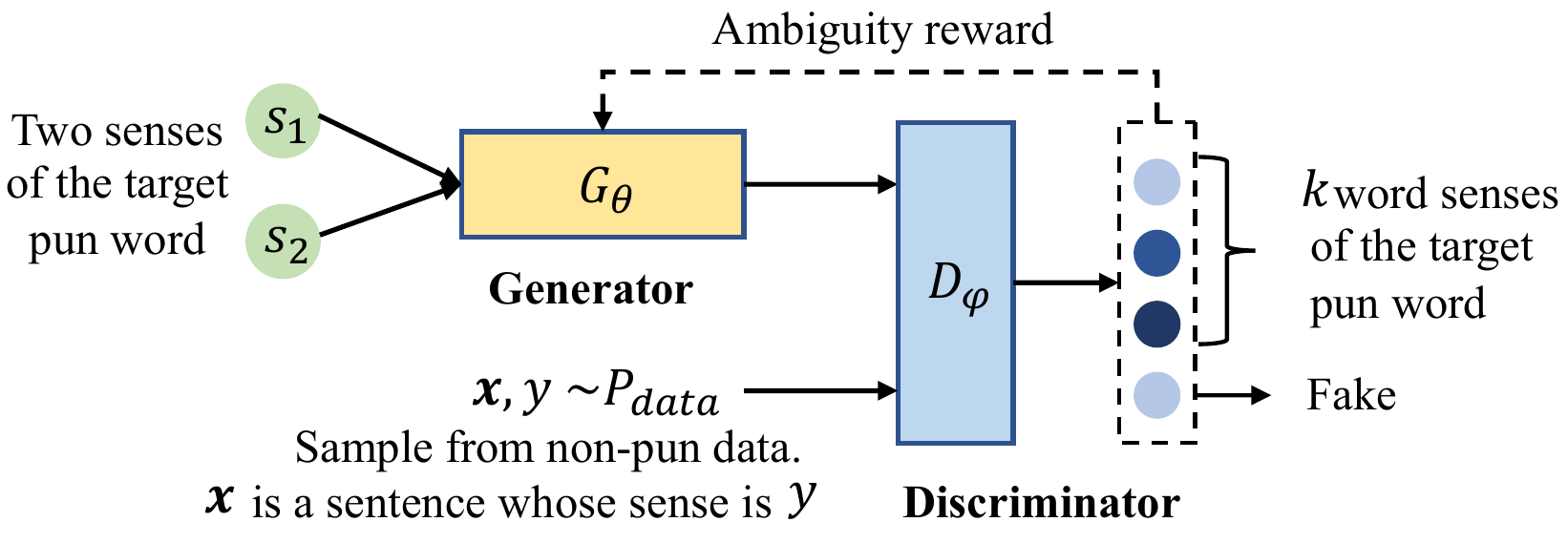}
    %\vspace{-0.2in}
	\caption{The proposed Pun-GAN framework.}\label{fig:model}
    % \vspace{-0.20in}
\end{figure}

Based on the above motivations, we introduce Generative Adversarial Net~\cite{newgan} into pun generation task. Specifically, the generator can be any model that is able to generate a pun sentence containing a given word with two specific senses.
% And the discriminator is a word sense classifier to distinguish the real non-pun sentences and generated pun sentences.
The discriminator is a word sense classifier to classify the real sentence to its correct word sense label and classify a generated pun sentence to a fake label.
With such a framework, the discriminator can provide a well-designed ambiguity reward to the generator, thus encouraging the ambiguity of the generated sentence via reinforcement learning (RL) to achieve the goal of punning, without using any pun corpus.

% Note that the generator is trained on non-pun corpus, and the discriminator is pre-trained only on sense-labeled corpus.

Evaluation of the pun generation is also challenging.
We conduct both automatic and human evaluations. The results show that the proposed Pun-GAN can generate a higher quality of pun sentence, especially in ambiguity and diversity.

\section{Model}

% \subsection{Overview}

The sketch of the proposed Pun-GAN is depicted in Figure~\ref{fig:model}.
It consists of a pun generator $G_\theta$ and a word sense discriminator $D_\phi$.
The following sections will elaborate on the architecture of Pun-GAN and its training algorithm.

\subsection{Model Structure}

\subsubsection{Generator}
Given two senses $(s_1, s_2)$ of a target word $w$, the generator $G_\theta$ aims to output a sentence $\bm x$ which not only contains the target word $w$ but also express the two corresponding meanings.
Considering the simplicity of the model and the ease of training, we adopt the neural constrained language model of ~\citet{Neural18Yu} as the generator.
Due to space constraints, we strongly recommend that readers refer to the original paper for details.
Compared with traditional neural language model, the main difference is that the generated words at each timestep should have the maximum sum of two probabilities which are calculated with $s_1$ and $s_2$ as input, \textit{respectively}.
% Compared to traditional neural language model, the main difference is that the generated words at each timestep should have the biggest sum of one probability calculated by taking $s_1$ as input and another probability calculated by taking $s_2$ as input.
Formally, the generation probability over the entire vocabulary at $t$-th timestep is calculated as
% \begin{equation}
%     G_\theta (x_t|\bm x_{<t}) = \mathrm{softmax}(Wh_1^t) + \mathrm{softmax}(Wh_2^t)
% \end{equation}
% where $h_1^t$ and $h_2^t$ are the hidden states of $t$-th decoding step when taking $s_1$ and $s_2$ as input, respectively. And $W$ is a trainable matrix.
\begin{equation}
    % G_\theta (\bm x_t|\bm x_{<t}) = \mathrm{softmax}(Wh_1^t) + \mathrm{softmax}(Wh_2^t)
    G_\theta (x_t|\bm x_{<t}) = f(Wh^1_t+b) + f(Wh^2_t+b)
\end{equation}
where $h^1_t$ ($h^2_t$) is the hidden state of $t$-th step when taking $s_1$ ($s_2$) as input, $f$ is the softmax function, and $\bm x_{<t}$ is the preceding $t-1$ words.

Therefore, the generation probability of the whole sentence $\bm x$ is formulated as

\begin{equation} \label{eq:joint}
    G_\theta (\bm x|s_1, s_2) = \prod_t G_\theta (x_t|\bm x_{<t})
\end{equation}

To give a warm start to the generator, we pre-train it using the same general training corpus in the original paper.

\subsubsection{Discriminator}
% The discriminator $D_\phi$ is trained to distinguish generated pun sentence and real non-pun sentence.
The discriminator is extended from the word sense disambiguation models~\cite{bilstm,Luo18WSD2,Luo18WSD1}.
Assuming the pun word $w$ in sentence $\bm x$ has $k$ word senses, we add a new ``generated'' class.
Then, the discriminator is designed to produce a probability distribution over $k + 1$ classes, which is computed as
% For each pun word $w$, suppose it have $k$ word senses. The discriminator is designed to produces a probability distribution over $k + 1$ classes.
% Following ~\citet{bilstm}, we adopt a bidirectional LSTM to extract the context vector $c$. Thus the probability distribution is
\begin{equation}
    D_\phi(y|\bm x) = \mathrm{softmax}(U_wc + b')
\end{equation}
where $c$ is the context vector from a bi-directional LSTM when taking $\bm x$ as input, $U_w$ is a word-specific parameter and $y$ is the target label.

Therefore, $D_\phi\big(y=i|x, i\in\{1,...,k\}\big)$ denotes the probability that it belongs to the real $i$-th word sense, while $D_\phi(y=k+1|x)$ denotes the probability that it is produced by a pun generator.
% Indeed, the discriminator is somewhat like a word sense disambiguation (WSD) model~\cite{bilstm}, except that we add one more class (label) for each word.
% Following previous work about WSD, the discriminator is a bidirectional LSTM model with a word-specific softmax layer over $(1, ..., k)$ word senses and a sigmoid layer over the $(k+1)$-th class. 

% \begin{equation}
%     \mathcal{J}(\phi) = \nabla_{\theta}{\rm log}\Big(p\big(\bm x| S(\bm x,v_y;\theta),v_x;\theta\big)\Big)
% \end{equation}

\subsection{Training}
We follow the training techniques of \citet{GAN} which applys GAN to semi-supervised learning. For real sentence $\bm x$, if it is sense labeled, $D_\phi$ should classify $\bm x$ to its correct word sense label $y$, otherwise $D_\phi$ should classify $\bm x$ to anyone of the $k$ labels. For generated sentence $\bm x$, $D_\phi$ should classify $\bm x$ to the $(k+1)$-th generated label.
% For discriminator, $D_\phi\big(y=i|x, i\in\{1,...,k\}\big)$ denotes the the probability that it belongs to the real $i$-th word sense, while $D_\phi(y=k+1|x)$ denotes the probability that it is generated by pun generators.
% What's more, in order to classify generated sentence and real true sentence, we add 
% The main difference is that the generated pun sentence is suitable for two specified senses of a homographic word, while the real non-pun sentence can be classified to only on word sense. 
Thus, the training objective of the discriminator is to minimize:
\begin{equation} \label{eq:E}
% \small
\begin{aligned}
\mathcal{J}(\phi) = & - \mathbb{E}_{\bm x,y \sim p_{\text{data}}(\bm x,y)}\log p_\phi(y|\bm x) \\
& - \mathbb{E}_{\bm x \sim p_{\text{data}}(\bm x)}\log p_\phi(y<k+1|\bm x) \\
% & - \mathbb{E}_{\bm x, s_1, s_2 \sim G_\theta}\log \big(p_\phi(s_1 | \bm x) p_\phi(s_2 | \bm x) \big)\\
% & - \mathbb{E}_{\bm x, s_1, s_2 \sim G_\theta}[ \log p_\phi(y=s_1 | \bm x) + \log p_\phi(y=s_2 | \bm x) ] \\
& - \mathbb{E}_{\bm x \sim G_\theta} \log p_\phi(y=k+1| \bm x)
\end{aligned}
\end{equation}
where $p_{\text{data}}$ denotes the sentence which only supports one word sense.
% where $p_{\text{data}}(\bm x, y)$ denotes the sentence which is labeled with one word sense and $p_{\text{data}}(\bm x)$ denotes the unlabeled sentence which has one word sense.

% To encourage the generator to produce pun text, the discriminator is required to assign higher reward to the ambiguous pun text which can be understood in two meanings simultaneously, and assign a lower reward to the real non-pun text which only expresses one meaning. Suppose $(p_1, ..., p_k)$ is probability distribution generated by the discriminator. For pun sentence, the probability of the target two senses should not only count on the most,  but also have a small gap.
% % That is to say, the bigger $p_{s_1} + p_{s_2}$ is and the smaller $|p_{s_1} - p_{s_2}|$ is, the sentence is more likely to a pun sentence.
% Thus, the reward is defined as
% \begin{equation}
%     r =  \frac {p_{s_1} + p_{s_2}} {|p_{s_1} - p_{s_2}| + 1}
% \end{equation}

To encourage the generator to produce pun text, the discriminator is required to assign a higher reward to the ambiguous pun text which can be interpreted as two meanings simultaneously. For pun sentence, the probability of the target two sense $D_\phi({s_1|\bm x})$ and $D_\phi(s_2|\bm x)$ should not only \textit{have a small gap}, but also \textit{account for the most}. For example, (0.1, 0.5, 0.4) and (0.1, 0.8, 0.1) are two probability distributions outputted from $D_\phi$. The former is more likely to be a pun with the second (0.5) and third (0.4) meaning, while the latter is mostly a generic single sense sentence with the second meaning (0.8).
Based on the above observations, the reward is designed as
% Suppose the discriminator can generate a distribution over all word senses $(1, ..., k)$, 
\begin{equation} \label{eq:reward}
    r =  \frac {D_\phi({s_1|\bm x}) + D_\phi(s_2|\bm x)} {|D_\phi(s_1|\bm x) -D_\phi(s_2|\bm x)| + 1}
\end{equation}
where $1$ is a coefficient that avoids the denominator being zero.
% where $p_{s_1} + p_{s_2}$ measures the probability of the top-2 target sense, and $p_{s_1} - p_{s_2}$ can measures how closely the two 

Then, the goal of generator training is to minimize the negative expected reward.
\begin{equation} 
\label{eq:expect}
% 期望+函数的形式
% \mathcal{L}(\theta) = - \mathbb{E}_{\bm{y} \sim p_{\theta}}[r(\bm{y})] 
% 值加求和的形式
% \small
\mathcal{L}(\theta) =  - \sum_{k} r^{(k)} G_\theta(\bm x^{(k)}|s_1, s_2)
\end{equation}
where $\bm{x}^{(k)}$ is the $k$-th sampled sequence, $r^{(k)}$ is the reward of $\bm{x}^{(k)}$.

% Finally, we adopt policy gradient~\cite{williams1992reinforcement}, one of the reinforcement learning methods, to train the generator based on Eq~\ref{eq:expect}.
By means of policy gradient method~\cite{williams1992reinforcement}, for each pair of senses $(s_1, s_2)$, the expected gradient of Eq.~\ref{eq:expect} can be approximated as:
\begin{equation} \label{eq:loss_RL}
\nabla_{\theta}\mathcal{L}(\theta) \simeq - \frac{1}{K}\sum_{k=1}^K r^{(k)}\nabla_{\theta}{\rm log}\big(G_{\theta}(\bm{x}^{(k)})\big) 
\end{equation}
where $K$ is the sample size.

Similar to other GANs~\cite{GAN,SeqGAN}, the generator and discriminator are trained alternatively.

\section{Experiment}

\subsection{Dataset}
\quad \textbf{Training Dataset:}
To keep in line with previous work~\cite{Neural18Yu}, we use a generic corpus -- English Wikipedia to train Pun-GAN.
For generator, we first tag each word in the English Wikipedia corpus with \textit{one} word sense using an unsupervised WSD tool\footnote{\url{https://github.com/alvations/pywsd}}.
Then we use the 2,595K tagged corpus to pre-train our generator.
For discriminator, we use several types of data for training: 1) SemCor~\cite{Luo18WSD2,Luo18WSD1} which is a manually annotated corpus for WSD, consisting of 226K sense annotations\footnote{The reason why we don't use SemCor to train generator is that this dataset is too small for training a language model.} (first part in Eq.\ref{eq:E}); 2) Wikipedia corpus as unlabeled corpus (second part in Eq.\ref{eq:E}); 3) Generated puns (third part in Eq.\ref{eq:E}).
% We split 2,595K  sentences for training and other 742K sentences for development.

\textbf{Evaluation Dataset:}
We use the pun dataset from SemEval 2017 task7~\cite{PunDataset} for evaluation. The dataset consists of 1274 human-written puns where target pun words are annotated with two word senses. During testing, we extract the word sense pair as the input of our model.

\subsection{Experimental Setting}
The generator is the same as~\citet{Neural18Yu}.
The discriminator is a single-layer bi-directional LSTM with hidden size 128.
We randomly initialize word embeddings with the dimension size of 300.
The sample size K is set as 32.
Batch size is 32 and learning rate is 0.001.
 The optimization algorithm is SGD.
Before adversarial training, we pre-train the generator for 5 epochs and pre-train the discriminator for 4 epochs. In adversarial training, the generator is trained every 1 step and the discriminator is trained every 5 steps.

\subsection{Baselines}
% zhiwei & shunyao
% All baseline models are based on the same sequence-to-sequence model with ours.
We compare with the following systems:

\textbf{LM}~\cite{RNNLM}: It is a normal recurrent neural language model which takes the target pun word as input. 

\textbf{CLM}~\cite{ConstrainedLM}: It is a constrained language model which guarantees that a pre-given word will appear in the generated sequence.

\textbf{CLM+JD}~\cite{Neural18Yu}: It is a state-of-the-art model for pun generation which extends a constrained language model by jointly decoding conditioned on two word senses. 

% \textbf{CLM+JD+RL}: We augment the CLM+JD model with reinforcement learning. Indeed, it is an ablated version of our model via fixing the discriminator after pre-training.
% \textbf{Pun-GAN}$_{\mathrm{fix} D}$: It is an ablated version of our model via fixing the discriminator after pre-training.

\subsection{Evaluation Metrics}
\quad \textbf{Automatic evaluation:}
% 加上Unusualness
We use two metrics to automatically evaluate the creativeness of the generated puns in terms of unusualness and diversity. Following \citet{metric_unusualness} and \citet{Pun18Percy}\footnote{~\citet{Pun18Percy} is a contemporaneous work.},
% we use a normalized measure to compute unusualness: we take the log-probability of the model to be evaluated and subtract the log-probability of a baseline (LM).
the unusualness is measured by subtracting the log-probability of training sentences from the log-probability of generated pun sentences.
% We subtract the log-probability of a baseline model from log-probability of the model to be evaluated 
Following~\citet{Neural18Yu}, the diversity is measured by the ratio of distinct unigrams (Dist-1) and bigrams (Dist-2) in generated sentences. 
% Following previous work~\cite{Neural18Yu,Pun18Percy}~\footnote{~\citet{Pun18Percy} is a contemporaneous work.}, we adopt two metrics to evaluate the generated pun sentences in terms of unusualness and diversity.

\textbf{Human evaluation:} Three annotators score the randomly sampled 100 outputs of different systems from 1 to 5 in terms of three criteria. Ambiguity evaluates how likely the sentence is a pun. Fluency measures whether the sentence is fluent. Overall is a comprehensive metric.

\subsection{Results} \label{sec:results}
% zhiwei

\begin{table}[t]
\centering
\footnotesize
\setlength{\tabcolsep}{3pt}
\begin{tabular}{l | c c c c}
\toprule
\textbf{Model} & \textbf{Unusualness} & \textbf{Dist-1} & \textbf{Dist-2}\\ 
\midrule
% % 1274 sentences
% LM~\cite{RNNLM} & - & 6.8 & 15.4   \\
% CLM~\cite{ConstrainedLM} & 45.0 & 8.3 & 17.9   \\
% CLM + JD~\cite{Neural18Yu} & 4.9 & 8.8 & 19.8 \\
% % CLM + JD + RL &  & \bf 46.6 & 9.7 & 23.2   \\
% \midrule
% Pun-GAN & \bf 50.4 & \bf 11.3 & \bf 26.2  \\
% \midrule
% Human & 137.8 & 27.9 & 73.5 \\
% 1274 sentences
LM~\cite{RNNLM} & - & 6.8 & 15.4   \\
CLM~\cite{ConstrainedLM} & 0.45 & 8.3 & 17.9   \\
CLM+JD~\cite{Neural18Yu} & 0.05 & 8.8 & 19.8 \\
% CLM + JD + RL &  & \bf 46.6 & 9.7 & 23.2   \\
\midrule
Pun-GAN & \bf 0.50 & \bf 11.3 & \bf 26.2  \\
\midrule
Human & 1.38 & 27.9 & 73.5 \\
\bottomrule
\end{tabular}
%\vspace{-0.05in}
\caption{Automatic evaluation results.} \label{tb:auto}
%\vspace{-0.1in}
\end{table}

\begin{table}[t]
\centering
\footnotesize
\setlength{\tabcolsep}{2pt}
\begin{tabular}{l | c c c c}
\toprule
\textbf{Model} & \textbf{Ambiguity} & \textbf{Fluency} & \textbf{Overall}\\ 
\midrule
LM~\cite{RNNLM} & 1.6 & 3.1 & 2.5  \\
CLM~\cite{ConstrainedLM} & 2.0 & 2.1 & 2.0  \\
CLM+JD~\cite{Neural18Yu} & 3.4 & 3.6 & 3.5   \\
% CLM + JD + RL & 3.4 & \bf 3.7 & 3.6 \\
\midrule
Pun-GAN & \bf 3.9 & \bf 3.7 & \bf 3.8  \\
\midrule
Human & 4.3 & 4.6 & 4.5  \\
\bottomrule
\end{tabular}
%\vspace{-0.05in}
\caption{Human evaluation results.} \label{tb:human}
%\vspace{-0.15in}
\end{table}

Table \ref{tb:auto} and Table \ref{tb:human} show the results of automatic evaluation and human evaluation, respectively.
% % 合并成一段的写法
% We find that Pun-GAN performs the best scores in terms of ambiguity, fluency and diversity.
We find that:
1) Pun-GAN achieves the best ambiguity score. This is in line with our expectations that adversarial training can better achieve the aim of punning;
2) Compared with CLM+JD which is actually the same as our pre-trained generator, Pun-GAN has a large improvement in unusualness. We assume that it is because the discriminator can promote to generate more creative and unexpected sentences to some extent via adversarial training;
3) Pun-GAN can generate more diverse sentence with different tokens and words. This phenomenon accords with previous work of GANs~\cite{sentiGAN}.

In addition, Table~\ref{table:ab_test} shows the A/B tests between two the models. It shows that Pun-GAN can generate more vivid pun sentences compared with the previous best model CLM+JD. However, there still exists a big gap between generated puns and human-written puns.

% To conclude, both automatic evaluation and human evaluation show the effectiveness of the proposed Pun-GAN.
To conclude, both automatic evaluation and human evaluation show the effectiveness of the proposed Pun-GAN, especially in ambiguity and diversity of generated pun sentences.

\begin{table}[t]
\centering
\footnotesize
\setlength{\tabcolsep}{5pt}
\begin{tabular}{l | c c c c}
\toprule
\textbf{Model} & \textbf{Ours} & \textbf{No Pref.} & \textbf{Others}\\ 
\midrule
Pun-GAN vs CLM+JD & 57 & 19 & 24 \\
Pun-GAN vs Human & 8 & 13 & 79 \\

\bottomrule
\end{tabular}
%\vspace{-0.05in}
\caption{Results from human A/B testing of different pairs of models. Each cell indicates the times that a judge preferred one of the models or no preference of them among 100 sentences.} \label{table:ab_test}
\end{table}

\subsection{Ablation Study}
\begin{table}[t]
\centering
\footnotesize
\begin{tabular}{l | c c c c}
\toprule
\textbf{Model} & \textbf{Unusualness} & \textbf{Dist-1} & \textbf{Dist-2}\\ 
\midrule
Full Model & \bf 0.50 & \bf 11.3 & \bf 26.2  \\
- adversarial leaning & 0.46 & 10.9 & 25.1   \\
\bottomrule
\end{tabular}
%\vspace{-0.05in}
\caption{Ablation study.} \label{tb:ablation}
%\vspace{-0.15in}
\end{table}

In order to validate the effectiveness of adversarial learning, we fix the discriminator after pre-training. In this way, the Pun-GAN is degraded to a pure reinforcement learning framework. Table~\ref{tb:ablation} shows the results, from which we can conclude that adversarial leaning can help improve the creativeness of generated puns. We speculate that this is because adversary training makes generators and discriminators compete with each other, thus making progress together.

\begin{figure}[]
	\centering
	\includegraphics[width=1\columnwidth]{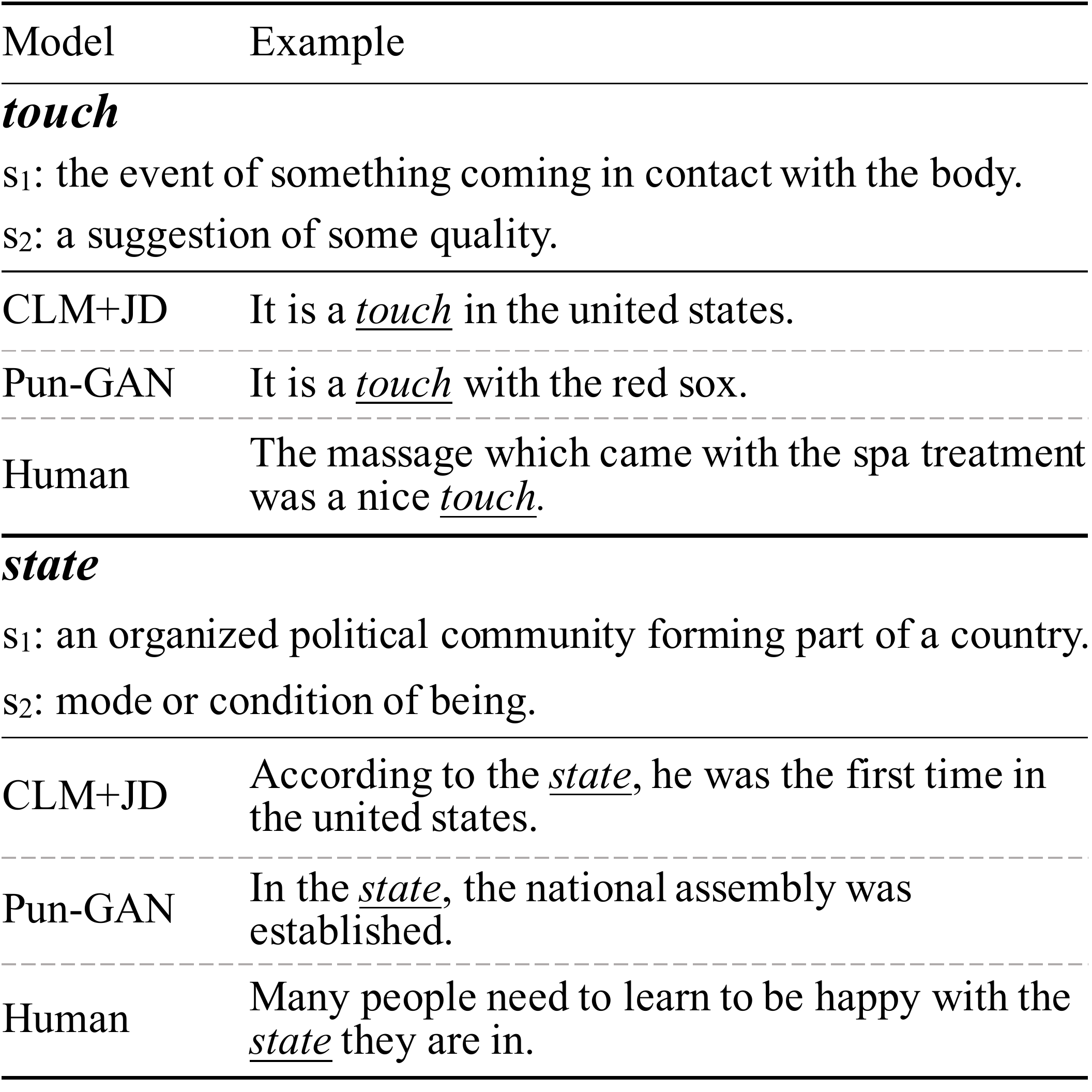}
	%\vspace{-0.2in}
	\caption{Example outputs of different models.}\label{fig:case}
	%\vspace{-0.10in}
\end{figure}

\subsection{Case Study}

Figure~\ref{fig:case} shows the randomly sampled examples of state-of-the-art model (CLM+JD) and Pun-GAN. Human-written puns are also given. It demonstrates that, compared with CLM+JD, Pun-GAN can generate puns which are closer to the funniness and creativeness of human-written puns. However, both CLM+JD and Pun-GAN may sometimes generate short sentences.
Since too short sentences lack sufficient context, they always tend to ambiguous. More analysis can be found in Section~\ref{sec:error}.

\subsection{Error Analysis} \label{sec:error}
\begin{figure}[t]
	\centering
    \includegraphics[width=0.6\columnwidth]{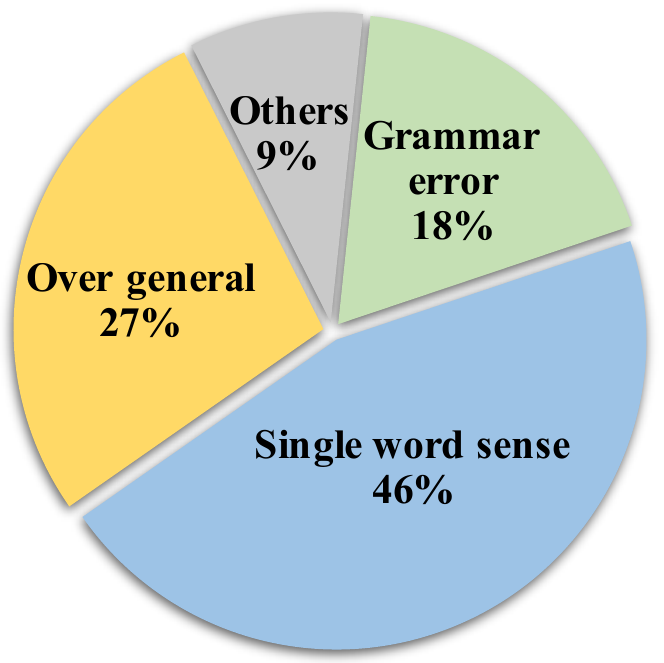}
    %\vspace{-0.05in}
	\caption{Pie chart of the error types.}\label{fig:error}
    %\vspace{-0.2in}
\end{figure}

We carefully analyze the generated results of Pun-GAN with low overall scores in human evaluation.
Fig~\ref{fig:error} shows the proportion of different error types. The most common type of error is generating a sentence which only supports a single word sense. This accords with our expectations since generating a sentence which can support two word senses without any labeled corpus is very hard. Another common type of error is generating over generic sentences. For example, ``\textit{It is a bank}''. In most instances, these generic sentences are always very short and they begin with a pronoun like ``\textit{It is}'' or ``\textit{He can}''. The reasons are two-fold. One is that these type of sentences can get a high generation probability since the generator is actually a language model. The other is these type of sentences can even get a not bad reward since they are indeed ambiguous.
Moreover, grammar error also accounts for about 1/5. We hypothesize that it is caused by the joint generation process in Eq.\ref{eq:joint}. 

 \section{Related Work}
Pun or humor generation is a very hard problem, which requires deep
natural language understanding. Early researches~\cite{pun13_template,pun13_template2} on pun generation are based on templates or rules for specific types of puns.  For example, ~\citet{pun13_template2}  propose a particular type
of pun: "\textit{I like my X like I like my Y, Z}", where \textit{X, Y} and \textit{Z} are words to be filled in based on word similarity and uncommonness. To improve the  creativity and flexibility of pun generation, ~\citet{Neural18Yu} propose an improved neural language model that is able to generate a sentence containing two word senses via using a joint decoding algorithm. ~\citet{Pun18Percy} proposes a retrieve-and-edit pipeline to generate a pun sentence. Another important point is that most of these methods doesn't need any pun corpus, since pun labeled training examples are really hard to require in practice. Strictly speaking, the generator of Pun-GAN can be any trainable pun generation model. For the sake of simplicity of the model and easiness to train, we adopt the neural approach of ~\citet{Neural18Yu} as the generator of our framework.

Our work is also in line with the recent progress of the application of GAN in text generation. The biggest challenge is that generating a text is a process of sampling in discrete space that is not differentiable, which makes it difficult to pass the gradients from discriminator to generator. To solve this non-differentiable, the most common is to train the generator via the policy gradient method of reinforcement learning \cite{SeqGAN, RankGAN, LeakGAN}. We borrow ideas from them and extends traditional binary discriminator to multi-class classifier like a word sense disambiguation model to better evaluate the ambiguity of the generated pun sentence.

\section{Conclusion and Future Work}
In this paper, we propose Pun-GAN: a generative adversarial network for pun generation.
It consists of a pun generator and a word sense discriminator, which unifies the task of pun generation and word sense disambiguation.
Even though Pun-GAN does not require any pun corpus, it can still enhance the ambiguity of sentence produced by the generator via the reward from the discriminator to achieve the goal of punning.
Pun-GAN is generic and flexible, and may be extended to other constrained text generation tasks in future work.

% \section*{Acknowledgments}
\section*{Acknowledgments}
This paper is supported by NSFC project 61772040 and 61751201.
The contact authors are Baobao Chang and Zhifang Sui.

\bibliography{emnlp-ijcnlp-2019}

\begin{thebibliography}{23}
\expandafter\ifx\csname natexlab\endcsname\relax\def\natexlab#1{#1}\fi

\bibitem[{Dey et~al.(2018)Dey, Juefei{-}Xu, Boddeti, and Savvides}]{RankGAN}
Rahul Dey, Felix Juefei{-}Xu, Vishnu~Naresh Boddeti, and Marios Savvides. 2018.
\newblock Rankgan: {A} maximum margin ranking {GAN} for generating faces.
\newblock \emph{CoRR}, abs/1812.08196.

\bibitem[{Goodfellow et~al.(2014)Goodfellow, Pouget{-}Abadie, Mirza, Xu,
  Warde{-}Farley, Ozair, Courville, and Bengio}]{newgan}
Ian~J. Goodfellow, Jean Pouget{-}Abadie, Mehdi Mirza, Bing Xu, David
  Warde{-}Farley, Sherjil Ozair, Aaron~C. Courville, and Yoshua Bengio. 2014.
\newblock Generative adversarial nets.
\newblock In \emph{Advances in Neural Information Processing Systems 27: Annual
  Conference on Neural Information Processing Systems 2014}, pages 2672--2680.

\bibitem[{Guo et~al.(2018)Guo, Lu, Cai, Zhang, Yu, and Wang}]{LeakGAN}
Jiaxian Guo, Sidi Lu, Han Cai, Weinan Zhang, Yong Yu, and Jun Wang. 2018.
\newblock Long text generation via adversarial training with leaked
  information.
\newblock In \emph{Proceedings of the Thirty-Second {AAAI} Conference on
  Artificial Intelligence, (AAAI-18)}.

\bibitem[{He et~al.(2019)He, Peng, and Liang}]{Pun18Percy}
He~He, Nanyun Peng, and Percy Liang. 2019.
\newblock Pun generation with surprise.
\newblock \emph{CoRR}, abs/1904.06828.

\bibitem[{Hong and Ong(2009)}]{Hong2009}
Bryan~Anthony Hong and Ethel Ong. 2009.
\newblock Automatically extracting word relationships as templates for pun
  generation.
\newblock In \emph{Proceedings of the Workshop on Computational Approaches to
  Linguistic Creativity}, CALC '09, pages 24--31.

\bibitem[{K{\aa}geb{\"a}ck and Salomonsson(2016)}]{bilstm}
Mikael K{\aa}geb{\"a}ck and Hans Salomonsson. 2016.
\newblock Word sense disambiguation using a bidirectional lstm.
\newblock \emph{arXiv preprint arXiv:1606.03568}.

\bibitem[{Luo et~al.(2018{\natexlab{a}})Luo, Liu, He, Xia, Sui, and
  Chang}]{Luo18WSD2}
Fuli Luo, Tianyu Liu, Zexue He, Qiaolin Xia, Zhifang Sui, and Baobao Chang.
  2018{\natexlab{a}}.
\newblock Leveraging gloss knowledge in neural word sense disambiguation by
  hierarchical co-attention.
\newblock In \emph{Proceedings of the 2018 Conference on Empirical Methods in
  Natural Language Processing, 2018}.

\bibitem[{Luo et~al.(2018{\natexlab{b}})Luo, Liu, Xia, Chang, and
  Sui}]{Luo18WSD1}
Fuli Luo, Tianyu Liu, Qiaolin Xia, Baobao Chang, and Zhifang Sui.
  2018{\natexlab{b}}.
\newblock Incorporating glosses into neural word sense disambiguation.
\newblock In \emph{Proceedings of the 56th Annual Meeting of the Association
  for Computational Linguistics, {ACL} 2018}.

\bibitem[{Mikolov et~al.(2010)Mikolov, Karafi{\'{a}}t, Burget, Cernock{\'{y}},
  and Khudanpur}]{RNNLM}
Tomas Mikolov, Martin Karafi{\'{a}}t, Luk{\'{a}}s Burget, Jan Cernock{\'{y}},
  and Sanjeev Khudanpur. 2010.
\newblock Recurrent neural network based language model.
\newblock In \emph{{INTERSPEECH} 2010, 11th Annual Conference of the
  International Speech Communication Association, 2010}, pages 1045--1048.

\bibitem[{Miller and Gurevych(2015)}]{MillerG15}
Tristan Miller and Iryna Gurevych. 2015.
\newblock Automatic disambiguation of english puns.
\newblock In \emph{Proceedings of the 53rd Annual Meeting of the Association
  for Computational Linguistics and the 7th International Joint Conference on
  Natural Language Processing of the Asian Federation of Natural Language
  Processing, {ACL} 2015, Volume 1: Long Papers}, pages 719--729.

\bibitem[{Miller et~al.(2017)Miller, Hempelmann, and Gurevych}]{PunDataset}
Tristan Miller, Christian Hempelmann, and Iryna Gurevych. 2017.
\newblock Semeval-2017 task 7: Detection and interpretation of english puns.
\newblock In \emph{Proceedings of the 11th International Workshop on Semantic
  Evaluation, SemEval@ACL 2017}, pages 58--68.

\bibitem[{Mou et~al.(2015)Mou, Yan, Li, Zhang, and Jin}]{ConstrainedLM}
Lili Mou, Rui Yan, Ge~Li, Lu~Zhang, and Zhi Jin. 2015.
\newblock Backbone language modeling for constrained natural language
  generation.
\newblock \emph{CoRR}, abs/1512.06612.

\bibitem[{Pal and Saha(2015)}]{WSD_survey}
Alok~Ranjan Pal and Diganta Saha. 2015.
\newblock Word sense disambiguation: a survey.
\newblock \emph{CoRR}, abs/1508.01346.

\bibitem[{Pauls and Klein(2012)}]{metric_unusualness}
Adam Pauls and Dan Klein. 2012.
\newblock Large-scale syntactic language modeling with treelets.
\newblock In \emph{The 50th Annual Meeting of the Association for Computational
  Linguistics, Proceedings of the Conference, July 8-14, 2012, Jeju Island,
  Korea - Volume 1: Long Papers}, pages 959--968. The Association for Computer
  Linguistics.

\bibitem[{Paulus et~al.(2017)Paulus, Xiong, and
  Socher}]{paulus2017RL_for_summarization}
Romain Paulus, Caiming Xiong, and Richard Socher. 2017.
\newblock A deep reinforced model for abstractive summarization.
\newblock In \emph{Proceedings of the International Conference on Learning
  Representations, {ICLR} 2017}.

\bibitem[{Petrovic and Matthews(2013)}]{pun13_template2}
Sasa Petrovic and David Matthews. 2013.
\newblock Unsupervised joke generation from big data.
\newblock In \emph{Proceedings of the 51st Annual Meeting of the Association
  for Computational Linguistics, {ACL} 2013, Volume 2: Short Papers}, pages
  228--232.

\bibitem[{Raganato et~al.(2017)Raganato, Bovi, and Navigli}]{Raganato2017}
Alessandro Raganato, Claudio~Delli Bovi, and Roberto Navigli. 2017.
\newblock Neural sequence learning models for word sense disambiguation.
\newblock In \emph{Conference on Empirical Methods in Natural Language
  Processing (EMNLP)}.

\bibitem[{Salimans et~al.(2016)Salimans, Goodfellow, Zaremba, Cheung, Radford,
  and Chen}]{GAN}
Tim Salimans, Ian~J. Goodfellow, Wojciech Zaremba, Vicki Cheung, Alec Radford,
  and Xi~Chen. 2016.
\newblock Improved techniques for training gans.
\newblock In \emph{Advances in Neural Information Processing Systems 29: Annual
  Conference on Neural Information Processing Systems 2016}, pages 2226--2234.

\bibitem[{Valitutti et~al.(2013)Valitutti, Toivonen, Doucet, and
  Toivanen}]{pun13_template}
Alessandro Valitutti, Hannu Toivonen, Antoine Doucet, and Jukka~M. Toivanen.
  2013.
\newblock ``let everything turn well in your wife'': Generation of adult humor
  using lexical constraints.
\newblock In \emph{Proceedings of the 51st Annual Meeting of the Association
  for Computational Linguistics, {ACL} 2013, Volume 2: Short Papers}, pages
  243--248.

\bibitem[{Wang and Wan(2018)}]{sentiGAN}
Ke~Wang and Xiaojun Wan. 2018.
\newblock Sentigan: Generating sentimental texts via mixture adversarial
  networks.
\newblock In \emph{Proceedings of the Twenty-Seventh International Joint
  Conference on Artificial Intelligence, {IJCAI} 2018}, pages 4446--4452.

\bibitem[{Williams(1992)}]{williams1992reinforcement}
Ronald~J. Williams. 1992.
\newblock Simple statistical gradient-following algorithms for connectionist
  reinforcement learning.
\newblock In \emph{Machine Learning}.

\bibitem[{Yu et~al.(2017)Yu, Zhang, Wang, and Yu}]{SeqGAN}
Lantao Yu, Weinan Zhang, Jun Wang, and Yong Yu. 2017.
\newblock Seqgan: Sequence generative adversarial nets with policy gradient.
\newblock In \emph{Proceedings of the Thirty-First {AAAI} Conference on
  Artificial Intelligence}, pages 2852--2858.

\bibitem[{Yu et~al.(2018)Yu, Tan, and Wan}]{Neural18Yu}
Zhiwei Yu, Jiwei Tan, and Xiaojun Wan. 2018.
\newblock A neural approach to pun generation.
\newblock In \emph{Proceedings of the 56th Annual Meeting of the Association
  for Computational Linguistics, {ACL} 2018}.

\end{thebibliography}
\bibliographystyle{acl_natbib}

% \appendix
% \section{Supplemental Material}
% \label{sec:supplemental}

\nocite{RankGAN}
\nocite{LeakGAN}
\nocite{paulus2017RL_for_summarization}
\nocite{Raganato2017}
\nocite{Pun18Percy}
\nocite{pun13_template2}

\end{document}